\documentclass[11pt]{article}

\usepackage[preprint]{acl}

\usepackage{times}
\usepackage{latexsym}
\usepackage[T1]{fontenc}
\usepackage[utf8]{inputenc}
\usepackage{microtype}
\usepackage{inconsolata}
\usepackage{graphicx}
\usepackage{booktabs}
\usepackage{multirow}
\usepackage{amsmath}
\usepackage{array}
\usepackage{xcolor}
\usepackage{url}
\usepackage{colortbl}
\usepackage{enumitem}
\usepackage{tikz}
\usetikzlibrary{positioning,arrows.meta,calc,fit,backgrounds,shapes.misc}
\usepackage[most]{tcolorbox}

\definecolor{acc}{HTML}{1F4E79}      
\definecolor{accsoft}{HTML}{E8EEF5}  
\definecolor{hum}{HTML}{7A4E00}      
\definecolor{humsoft}{HTML}{F6EEDD}
\definecolor{run}{HTML}{2E6B3A}      
\definecolor{runsoft}{HTML}{E5F0E6}
\definecolor{warn}{HTML}{A4471C}     
\definecolor{warnsoft}{HTML}{F8E6DC}
\definecolor{rule}{HTML}{9AA5B1}     
\definecolor{zebra}{HTML}{F4F6F8}    

\newcommand{\statustag}[3]{{\setlength{\fboxsep}{1.2pt}\colorbox{#2}{\textcolor{#1}{\scriptsize\sffamily #3}}}}
\newcommand{\accepted}{\statustag{run}{runsoft}{accepted}}
\newcommand{\candidate}{\statustag{warn}{warnsoft}{candidate}}

\newtcolorbox{artefactbox}[1]{enhanced, breakable, colback=zebra, colframe=rule,
  boxrule=0.4pt, arc=1.5pt, left=4pt, right=4pt, top=3pt, bottom=3pt,
  fonttitle=\sffamily\scriptsize\bfseries, coltitle=acc, colbacktitle=accsoft,
  title={#1}, attach boxed title to top left={xshift=4pt,yshift=-2pt},
  boxed title style={boxrule=0.4pt, colframe=rule, arc=1pt}}

\title{Can an Agent Debug a Web Corpus?\\Agentic Auditing of Slovak FineWeb2}

\author{Marek Šuppa\textsuperscript{1,2} \quad Ivan Vykopal\textsuperscript{3,4} \quad Jaroslav Kopčan\textsuperscript{3} \\
\textsuperscript{1}Comenius University in Bratislava \quad \textsuperscript{2}Cisco Systems \\
\textsuperscript{3}Kempelen Institute of Intelligent Technologies (KInIT) \\
\textsuperscript{4}Brno University of Technology}

\newcommand{\vallabels}{255}
\newcommand{\valcellmin}{12}
\newcommand{\valcellmax}{21}
\newcommand{\valcellmed}{19}
\newcommand{\suiten}{9}
\newcommand{\nretained}{7}
\newcommand{\suitesecs}{0.7}
\newcommand{\suitedocs}{26{,}470{,}482}
\newcommand{\farmdocs}{1{,}408{,}437}

\newcommand{\invalidshare}{40.4}
\newcommand{\corpusM}{26.5}
\newcommand{\budget}{300}

\newcommand{\farmwordshare}{12.3}
\newcommand{\iaasentence}{}
\newcommand{\codingrulesentence}{ Labels were collected in two passes, and the criterion for unusable text was clarified between them: a fluent page that exists to place links or keywords counts as junk. 117 of the 255 labels were given before it and the annotator reviewed and kept them rather than judging those documents again, which matters most for the two slices where they dominate (12 of 14 and 11 of 17).}

\newcommand{\validagent}{6}

\newcommand{\thresholdaccept}{.5}
\newcommand{\valpmin}{.53}
\newcommand{\valpmax}{1.00}
\newcommand{\valunsup}{.07}

\newcommand{\valcontrol}{.08}
\newcommand{\valcontrolhi}{.25}
\newcommand{\valcontroln}{25}

\newcommand{\runvalidmin}{2}
\newcommand{\runvalidmax}{3}
\newcommand{\rungroupsmin}{2}
\newcommand{\rungroupsmax}{3}
\newcommand{\basevalidmin}{2}
\newcommand{\basevalidbest}{4}
\newcommand{\basevalidbestname}{\textsc{LowConf}}
\newcommand{\poolagentvalid}{6}
\newcommand{\poolagentgroups}{3}
\newcommand{\poolbasevalid}{6}
\newcommand{\poolbasegroups}{4}

\newcommand{\runusdmin}{2.21}
\newcommand{\runusdmax}{2.84}
\newcommand{\baseusdmin}{1.03}
\newcommand{\baseusdmax}{1.22}
\newcommand{\poolbasecand}{21}
\newcommand{\basecandsum}{23}
\newcommand{\agentonlygroups}{G04}
\newcommand{\baseonlygroups}{G09, G10}
\newcommand{\gtwobaselines}{3}
\newcommand{\gtwobasepmin}{.57}
\newcommand{\gtwobasepmax}{.78}
\newcommand{\gfourpmin}{0.20}
\newcommand{\gfourpmax}{0.39}
\newcommand{\gnineprec}{1.00}
\newcommand{\gninen}{9}
\newcommand{\gninedocs}{1{,}717}
\newcommand{\ldadultsel}{language='slk' AND n\_bad\_words >= 4}
\newcommand{\ldadultpct}{0.30}
\newcommand{\anndrawn}{358}
\newcommand{\annlabelled}{255}
\newcommand{\annpassone}{235}
\newcommand{\annpasstwo}{138}
\newcommand{\annpasstwonew}{20}
\newcommand{\annrecodelisted}{171}
\newcommand{\annbasecand}{12}
\newcommand{\annbasemin}{11}
\newcommand{\annbasemax}{12}
\newcommand{\anncontrol}{40}
\newcommand{\defectlang}{\texttt{cs}, \texttt{mix}, \texttt{other}}
\newcommand{\defectqual}{\texttt{junk}}
\newcommand{\annminhosts}{18}
\newcommand{\anntophost}{16}
\newcommand{\adjndecisions}{24}
\newcommand{\adjnclusters}{11}
\newcommand{\adjncandraw}{34}
\newcommand{\adjngrouprows}{10}
\newcommand{\adjnmemberrows}{14}
\newcommand{\adjnapprove}{11}
\newcommand{\adjnreject}{8}
\newcommand{\adjnunsure}{5}
\newcommand{\adjsplitgroup}{G02}
\newcommand{\adjsplitpmin}{.14}
\newcommand{\adjsplitpmax}{.81}

\newcommand{\farmpct}{5.32}
\newcommand{\farmpassall}{86.6}
\newcommand{\farmpassalldocs}{1{,}220{,}394}

\newcommand{\restpassall}{52.0}
\newcommand{\xlnconf}{12}
\newcommand{\xlnfail}{11}

\newcommand{\clsperk}{3.42}
\newcommand{\clsfull}{90{,}456}
\newcommand{\clstoken}{12{,}309}
\newcommand{\clsfullk}{90k}
\newcommand{\clstokenk}{12k}
\newcommand{\opususd}{8.17}
\newcommand{\opusturns}{147}
\newcommand{\opusratiomin}{2.9}
\newcommand{\opusratiomax}{3.7}
\newcommand{\replayn}{147}
\newcommand{\replayset}{144}
\newcommand{\replayexact}{139}
\newcommand{\replaysetpct}{98.0}
\newcommand{\replayexactpct}{94.6}
\newcommand{\replaysetdiff}{3}
\newcommand{\replayorderonly}{5}
\newcommand{\annbysource}{agent slices 122/175, \textsc{LowConf} 26/36, \textsc{LID-Disagree} 30/36, \textsc{HighDup} 28/36, \textsc{Random} 24/35, control 25/40}
\newcommand{\profbudget}{300}
\newcommand{\profbwmin}{15}
\newcommand{\profbwmax}{31}
\newcommand{\profoutside}{0}
\newcommand{\profslicedocs}{102{,}474}
\newcommand{\profslicepct}{0.39}
\newcommand{\profcoversall}{all three}

\begin{document}
\maketitle

\begin{abstract}
Multilingual pipelines such as FineWeb2 apply one recipe across a thousand languages, leaving the failure modes of individual languages unanticipated. We use an LLM once, as a corpus
\emph{debugger} rather than a classifier: a tool-using agent explores the corpus
through a read-only, budgeted API and compiles each finding into an executable test,
which a native speaker then validates. On the \corpusM{}M-document Slovak portion of
FineWeb2, three runs costing \$\runusdmin--\$\runusdmax{} each found seven findings over four phenomena. The largest is a machine-generated content farm of
\farmdocs{} documents (\farmpct\%) that passed the pipeline's filters,
\farmpassall\% of it passing even stricter thresholds. Blinded labels support \validagent{} of the \nretained{} findings
(defect rates \valpmin--\valpmax{} against \valcontrol{} in a random control) and do
not support one. The compiled suite scans the full corpus in \suitesecs\,s of
wall-clock time without a GPU, against \$\clstokenk--\$\clsfullk{} per LLM classification pass. The
best static sampling baseline validates more selectors than any agent run, at under half
the cost. The agent contributes exact prevalence, a precision estimate checked
against sampled documents, an account of why the pipeline missed each slice, and a
reusable test. We
release the framework, Slovak test suite, validation data and agent
trajectories.
\end{abstract}

\section{Introduction}
\label{sec:intro}

Pretraining corpora are now built by pipelines rather than by hand. FineWeb2
\citep{penedo2025fineweb2pipelinescale} derives more than a thousand
language-specific corpora from Common Crawl with one configurable recipe. That
uniformity is what makes such resources feasible at all. One pipeline, engineered
once, yields data for languages whose communities could never afford a bespoke
crawl. It is also a structural blind spot, as the FineWeb2 authors say
themselves. Their language identifier confuses closely related languages, and
they could not test each language individually, so they ask users to review the
filtering for the languages they care about. Their one mechanism against
cousin-language contamination, a high-affinity wordlist filter, is applied only to
languages whose measured contamination exceeded 10\%, and not to a mid-resource
language like Slovak with a large, closely related neighbour.

Reviewing a pipeline for one language is exactly the work that does not scale.
Its established form is the manual audit
\citep{kreutzer-etal-2022-quality,caswell-etal-2020-language}, which is
invaluable, and recent single-language audits still pay off
\citep{signoroni2026chinasdalsoch,tatariya-etal-2026-good}. But an audit is a
one-shot artefact. It describes a corpus at one point in time, does not compose into
machinery that can be re-run on the next crawl, and consumes expert attention
linearly. The alternative, running an LLM over the data as a classifier, turns a
curation problem into an inference bill and still explains nothing.

We explore a third option and use the LLM once, as a \emph{debugger}. Our auditor
is a tool-using agent that sees the corpus only through a small read-only
API for aggregate SQL, conditional sampling, text search, sentence-level language
identification, and defining and evaluating candidate
detectors. It is asked, in deliberately generic terms, to find systematic
defects and substantiate them. Its prompt contains no taxonomy of expected
problems, names no defect or neighbouring language, and describes the pipeline only by
its steps (language identification, quality heuristics, deduplication). Each output is a \emph{finding} with a description, a cheap executable
selector, a prevalence measured on the whole corpus, and a precision for human
validation to establish. The result is \emph{pytest for web
corpora}: a community contributes tests, and any future crawl or pipeline
revision is checked against them for the cost of a scan over metadata the pipeline
already computed. We instantiate it on Slovak\footnote{The framework, the adapter,
the Slovak test suite, the validation package and the agent trajectories are released
at \url{https://github.com/NaiveNeuron/agentic-corpus-debugger}.}, which has five million speakers and a large
closely related neighbour, and whose \corpusM{}M-document FineWeb2 slice is neither
high-resource enough to have received a dedicated audit nor small enough to read.

We frame an LLM, used once, as a corpus debugger whose output is a test rather than
a label or a data-selection policy. We build a reusable audit environment (DuckDB
over Parquet, eleven read-only budgeted tools, a finding schema) with a FineWeb2
adapter, and an agentic auditor that runs inside it under a fixed inspection budget
and a prompt that names neither a defect nor a neighbouring language. Around it sits a
three-stage workflow (discovery by the agent, validation by a native speaker,
deployment by a test runner with no model) in which validation can fail a finding.
It failed one of our seven findings, which comes out at a \valunsup{} defect rate and ships
flagged. Our audit of Slovak FineWeb2 yields seven findings over four phenomena with
exact corpus-wide prevalences and blinded human precision (Table~\ref{tab:findings}),
among them a \farmpct\% content farm that the pipeline's own language, duplication and
auxiliary-LID signals do not filter, and a \texttt{language\_score} invariant that
fails in \xlnfail{} of \xlnconf{} FineWeb2 configurations. A per-run budget-matched
comparison against four static selection strategies shows what the agent does and
does not buy. It validates \runvalidmin--\runvalidmax{} selectors per run against
\basevalidmin--\basevalidbest{} for the baselines, so the best baseline finds more than any agent run, and each
phenomenon it validated also lies within reach of a static strategy, although for
the adult spam we analysed that strategy's sample without running the model on it (\S\ref{sec:results}).
Its contribution is the measurement, the characterisation and the test. Finally, we release the Slovak
regression-test suite, the patcher that turns it into a drop list, and the agent
trajectories (Table~\ref{tab:release}). Every defect we report is a residual of an
already heavily filtered corpus, and several are visible only \emph{because} FineWeb2
publishes rich metadata.

We ask four research questions.
\begin{enumerate}[label={\sffamily\bfseries\color{acc}RQ\arabic*},leftmargin=2.6em,labelsep=0.5em,itemsep=1pt,topsep=3pt,parsep=0pt]
\item Can an LLM agent with read-only tools discover systematic defects that a
native speaker confirms?
\item Can these defects be compiled into cheap executable tests?
\item Do the agent's findings lie outside the slices that static audit
strategies select?
\item At a matched per-run inspection budget, does the agent surface more
human-validated defects than the same model reading a statically selected sample?
\end{enumerate}

\section{Related Work}
\label{sec:related}

\paragraph{Web corpora and their audits.}
FineWeb2 \citep{penedo2025fineweb2pipelinescale} extends the FineWeb recipe
\citep{penedo2024finewebdatasetsdecantingweb} to over a thousand languages using
GlotLID \citep{kargaran-etal-2023-glotlid} with a per-language confidence
threshold, Gopher- and FineWeb-style quality filters, and global per-language
MinHash deduplication with rehydration. Earlier multilingual web corpora
\citep{ortiz-suarez-etal-2020-monolingual,xue-etal-2021-mt5,
conneau-etal-2020-unsupervised,kudugunta2023madlad400multilingualdocumentlevellarge,
nguyen-etal-2024-culturax,burchell-etal-2025-expanded} share its architecture and
its exposure to the LID error that \citet{caswell-etal-2020-language} and
\citet{burchell-etal-2023-open} document. Audits of such corpora are an
established genre, covering 205 corpora by hand \citep{kreutzer-etal-2022-quality},
one English corpus in depth \citep{dodge-etal-2021-documenting}, non-English
Wikipedia \citep{tatariya-etal-2026-good} and Lombard
\citep{signoroni2026chinasdalsoch}. The WaC community studies the same
failure surface from the pipeline side \citep{jakubicek-etal-2020-current,
barbaresi-lejeune-2020-box,laippala-etal-2020-web}. We are downstream of these, asking how an audit
can be made repeatable and executable. For Slovak, cleaning practice is
established \citep{pikuliak-etal-2022-slovakbert} and Czech/Slovak is a known hard
case for language ID \citep{tichy2026chalis}. Sub-percent contamination still means
hundreds of millions of tokens \citep{blevins-zettlemoyer-2022-language}, though we
promise no downstream gains \citep{artetxe-etal-2022-corpus}.

\paragraph{Tests over data, and agents over data.}
CheckList \citep{ribeiro-etal-2020-beyond} argued that behavioural tests expose
model failures that aggregate accuracy hides. We make that argument for corpora,
and borrow the pattern of declarative data-quality assertions run on every build
\citep{DBLP:journals/pvldb/SchelterLSCBG18}, except that here the assertions must
come from people who know the language. LLM agents have been applied to
corpus-linguistic discovery \citep{yu2026agentdrivencorpuslinguisticsframework}
and to the curation loop, where \citet{kang2026generalistagentsautomatedata} find
that coding agents optimise data-selection policies but tend to tune variants of a
known policy. Ours optimises no downstream metric and outputs a test suite instead of
a policy. We do not claim that LLMs are good language classifiers. The agent
writes the tests that label the corpus and never labels it itself.

\section{The Audit Environment}
\label{sec:env}

\paragraph{Corpus view.}
The adapter exposes a corpus as two DuckDB relations over Parquet.
\texttt{corpus\_meta} has one row per document without text and serves exact
corpus-wide prevalence, while \texttt{corpus} is a deterministic 1-in-13 subsample
with text (2{,}034{,}894 documents) for whenever reading is needed. That split keeps
the environment interactive, because metadata aggregates over \corpusM{}M rows return in under a
second while scanning 100\,GB of text does not. The columns are the corpus's own
metadata (URL, crawl, date, the pipeline's \texttt{language} and
\texttt{language\_score}, \texttt{minhash\_cluster\_size}, the profanity-list count
\texttt{n\_bad\_words}, the GlotLID \texttt{top\_langs} distribution, and the
fastText, langdetect and Lingua predictions added by the
\texttt{ivykopal/fineweb2-slovak} release), plus language-neutral derived columns
(domain, size counts). Nothing in the schema names a language pair or a failure mode.

\paragraph{Tools.}
The agent acts through eleven tools, each listed with its budget cost in
Appendix~\ref{app:tools}. They cover aggregation, sampling and single-document
inspection, text search, host and crawl statistics, sentence-level language
identification, two tools that measure and sample what a candidate rule selects, and
\texttt{save\_finding}. Aggregation is free, whereas every document read costs one inspection from a fixed
budget, charged once per distinct id. Every call is logged to a JSONL trajectory,
so an investigation can be replayed.
The free aggregation tool did not stop a query from returning document text, for
example \texttt{substr(text, 1, 120)} grouped and counted, so text could in principle
be read without spending budget. We checked every logged aggregate call. In the three
protocol runs none returned document text: the queries that touch \texttt{text} return
counts only. The fourth run in Appendix~\ref{app:robust} returned text fragments in
three calls, the most frequent 100- and 120-character document prefixes and extracted
e-mail strings, and its inspection count therefore understates what it read. Later
versions of the environment refuse such queries.

\paragraph{Agent loop and prompt hygiene.}
The agent runs one long trajectory in which it orients on statistics, forms a
hypothesis, retrieves evidence and counter-evidence, refines or rejects it, measures
prevalence, synthesises a selector, checks its precision and records a finding. Its system prompt
(Appendix~\ref{app:prompt}) states the tools, the finding schema and the
epistemic standard (systematic, reproducibly retrievable, grounded in inspected
documents, quantified), and asks for calibrated precision estimates and
counter-evidence. Because it does not name any defect, failure mode or neighbouring
language, discovery stays attributable to the agent.
Appendix~\ref{app:integrity} verifies this mechanically and states exactly what the agent was told. We run it through the Claude Code CLI with only the audit tools and Claude Code's tool-schema lookup enabled, and no shell, file or web access.

\begin{figure}[t]
\centering\footnotesize
\begin{tikzpicture}[
  font=\footnotesize,
  stage/.style={draw=#1, fill=#1soft, line width=0.5pt, rounded corners=1.5pt,
    minimum height=1.05cm, text width=0.18\columnwidth, align=center,
    inner sep=2pt, font=\sffamily\scriptsize\bfseries, text=#1},
  body/.style={draw=#1, line width=0.4pt, rounded corners=1.5pt, fill=white,
    minimum height=1.05cm, text width=0.645\columnwidth, align=left,
    inner sep=3pt, font=\scriptsize},
  flow/.style={-{Stealth[length=4pt]}, line width=0.6pt, draw=acc},
  node distance=2.6mm and 1.2mm]
\newcommand{\lab}[1]{{\sffamily\scriptsize\color{black!60}#1}\ }
\node[stage=acc] (s1) {1.\\Discovery\\[1pt]\mdseries LLM agent};
\node[body=acc, right=of s1] (b1) {%
  \lab{Actor/cost} LLM agent, read-only API, $\leq$\budget{} inspections;
  \$\runusdmin--\$\runusdmax{} per run\\[1pt]
  \lab{Output} candidate findings: selector, exact prevalence, expected
  precision, evidence};
\node[stage=hum, below=of s1] (s2a) {2a.\\Labels\\[1pt]\mdseries human};
\node[body=hum, right=of s2a] (b2a) {%
  \lab{Actor/cost} native speaker, blinded sample; \vallabels{} labels\\[1pt]
  \lab{Output} defect rate with 95\% interval};
\node[stage=hum, below=of s2a, dashed] (s2b) {2b.\\Verdicts\\[1pt]\mdseries
  model$\,\to$\\\mdseries human};
\node[body=hum, right=of s2b] (b2b) {%
  \lab{Actor/cost} model \emph{proposes} from the tabulated labels; annotator
  \emph{adopts}\\[1pt]
  \lab{Output} status \texttt{accepted} or \texttt{candidate}};
\node[stage=run, below=of s2b] (s3) {3.\\Deployment\\[1pt]\mdseries test runner};
\node[body=run, right=of s3] (b3) {%
  \lab{Actor/cost} test runner without an LLM or GPU; \suitesecs\,s wall-clock over
  \corpusM{}M documents\\[1pt]
  \lab{Output} prevalence guards and invariants on each release; a drop list};
\draw[hum, line width=0.6pt] ([xshift=-1.5mm]s2a.north west) -- ++(-0.8mm,0)
  |- ([xshift=-2.3mm]s2b.south west) -- ++(0.8mm,0);
\node[rotate=90, font=\sffamily\scriptsize\bfseries, text=hum, anchor=south]
  at ($([xshift=-2.3mm]s2a.north west)!0.5!([xshift=-2.3mm]s2b.south west)$)
  {Validation};
\draw[flow] (s1) -- (s2a);
\draw[flow, draw=hum] (s2a) -- (s2b);
\draw[flow] (s2b) -- (s3);
\end{tikzpicture}
\caption{The three stages. Every label is human. Verdicts are proposed by a model from
those labels and adopted by the annotator, and only an adopted verdict assigns a
status. Deployment uses no model.}
\label{fig:stages}
\end{figure}
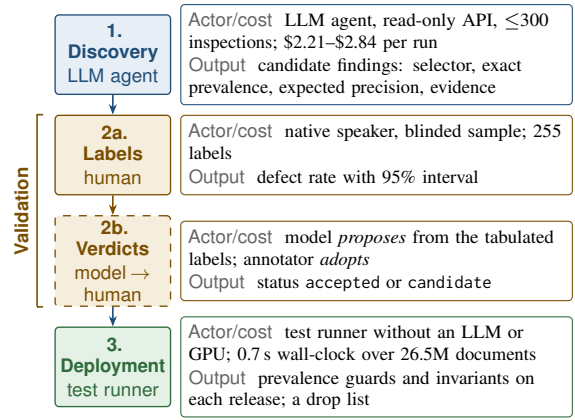

\paragraph{Discovery, validation, deployment.}
The workflow has three stages with different actors and costs
(Figure~\ref{fig:stages}). \emph{Discovery} is where the agent works, and it produces
candidates. In \emph{validation} a native speaker labels a blinded sample of each
candidate's documents, after which a model tabulates that evidence and proposes a verdict
per candidate group that the annotator adopts or not (\S\ref{sec:setup}).
\emph{Deployment} runs the resulting tests without any model. Each test is a YAML
file carrying its selector, an assertion about future corpora and its evidence
(Appendix~\ref{app:findings}), and a status: \texttt{accepted} when the blinded labels
support it and the adopted verdict approves it, \texttt{candidate} otherwise. A suite
is a directory of such files, executable against any corpus implementing the adapter
interface.

\paragraph{Release.}
The release (Table~\ref{tab:release}) is the artefact a second community would reuse:
the environment and runner, the prompt, the suite, and every input to the numbers in
this paper, from trajectories to verdicts. The paper's tables are generated from these
artefacts by script.

\begin{table}[t]
\centering\footnotesize
\setlength{\tabcolsep}{3pt}
\renewcommand{\arraystretch}{1.12}
\providecommand{\repogroup}[1]{\rowcolor{accsoft}\multicolumn{2}{@{\hspace{3pt}}l@{}}{\scriptsize\sffamily\bfseries\color{acc}#1}\\}
\begin{tabular}{@{}>{\raggedright\arraybackslash}p{0.36\columnwidth}
>{\raggedright\arraybackslash}p{0.59\columnwidth}@{}}
\toprule
\textbf{Path} & \textbf{Contents} \\
\midrule
\repogroup{CODE}
\texttt{src/corpusaudit/} & adapter (DuckDB over Parquet), the eleven tools as an MCP
server, test runner, patcher \\
\texttt{prompts/} & the agent and baseline prompts \\
\texttt{tests/} & the suite: 7 Slovak slice guards, 2 metadata invariants \\
\repogroup{DATA \& RUNS}
\texttt{out/runs/}, \texttt{out/baselines/} & trajectories (every tool call) and
baseline outputs \\
\texttt{out/annotation/} & blinded package, guidelines, both label passes \\
\texttt{out/adjudication/} & candidate groups, verdicts, how they were reached \\
\repogroup{ANALYSIS \& REPORTS}
\texttt{out/analysis/}, \texttt{out/REPORT.md} & every measurement, and a generated
report per finding \\
\repogroup{TOOLING}
\texttt{scripts/}, \texttt{selftest/} & regeneration and checks, the drop-list
writer; a synthetic-corpus self-test that needs no data \\
\bottomrule
\end{tabular}
\caption{Layout of the released repository.}
\label{tab:release}
\end{table}

\section{Experimental Setup}
\label{sec:setup}

\paragraph{Data.}
We audit the Slovak portion of FineWeb2 (\texttt{slk\_Latn}) as redistributed with
extra LID metadata by \texttt{ivykopal/fineweb2-slovak}. It holds \suitedocs{}
documents from 351{,}844 hosts and 96 Common Crawl snapshots (2013--2024), 438 words on
average. No document scores below 0.667, the \texttt{language\_score} threshold
FineWeb2 publishes for \texttt{slk\_Latn}. The auxiliary identifiers disagree with
the pipeline label on 3.12\% of documents.

\paragraph{Models and settings.}
All three agent runs and all four baselines use \texttt{claude-sonnet-5} through the
Claude Code CLI (version 2.1.231) at its default sampling settings, and we set no
temperature. Each agent run is capped at \budget{} document inspections and 250
turns, each baseline at 6 turns. We report inspections, tokens and USD for every run
and baseline. A fourth \texttt{claude-opus-5} run, outside this protocol,
is reported in Appendix~\ref{app:robust}.

\paragraph{Baselines.}
Each baseline is a \emph{document-selection strategy} that draws the same \budget{}
documents the agent could have inspected and shows them to the same model, without
tools or exploration. The model sees each document as its metadata and first 400
characters, whereas the agent can read whole documents, and the baseline
instructions drop the requirement that a finding be quantified and have no
prevalence field.
We then measure the prevalence of each proposed selector ourselves, as for the
agent. \textsc{Random} samples uniformly. \textsc{LowConf}
takes $\texttt{language\_score} < 0.8$, which is 0.49\% of the corpus,
\textsc{LID-Disagree} any auxiliary identifier disagreeing with the pipeline label
(3.12\%), and \textsc{HighDup} $\texttt{minhash\_cluster\_size} \geq 100$ (0.23\%).
The comparison changes two things at once. The agent chooses what to read, and it also has tools the baselines lack (aggregate queries, prevalence measurement,
targeted sampling, detector evaluation). It therefore measures the whole agentic
setup rather than adaptivity alone. For a fifth, profanity-ranked strategy we analyse only
the sample it would draw, the \profbudget{} text-subsample documents with the most
profanity-list matches, and we did not run the model on it.

\paragraph{Human validation.}
\label{par:human}
We drew \anndrawn{} documents for blinded annotation by a native speaker: 25 for each
of the 7 agent findings, \annbasemin--\annbasemax{} for each of \annbasecand{} sampled
baseline candidates, and \anncontrol{} uniform controls, excluding every document any
agent run inspected, shuffled and shown without provenance, URL or metadata. The
exclusion list was built from truncated tool logs, so one drawn document (a
code-switched Czech/Slovak item) had in fact been shown to an agent run by
\texttt{evaluate\_detector}. It received no label, so no reported number changes. Each
document gets a language label and a usefulness label on the scales in
Appendix~\ref{app:annotation}. A document is a \emph{defect} if its language label is
\defectlang{} or its usefulness label is \defectqual{}. \texttt{sk}, \texttt{sk-pred},
\texttt{good} and \texttt{noisy} are not defects. Slice precision is the human-judged
defect rate and the control gives the corpus base rate. This estimates precision, how often a selected document is a defect, not recall, how much of a phenomenon a selector catches.

Of the \anndrawn{} documents, \annlabelled{} carry a label, none of them
``uncertain'' (labelled/drawn by source: \annbysource{}). The shortfall is not a
per-selector choice: labelling ran in two passes with a clarified junk criterion
between them, and the documents neither pass reached were dropped
(Appendix~\ref{app:annotation}). Per-finding counts are in Table~\ref{tab:findings}.

\paragraph{Adjudication.}
The agent runs and the baselines saved \adjncandraw{} candidate files (9 agent and
25 baseline). They hold \nretained{} distinct agent selectors and \basecandsum{}
baseline selectors counted per baseline, \poolbasecand{} distinct across baselines,
since \textsc{Random} and \textsc{LowConf} share two. They were
clustered mechanically into \adjnclusters{} groups by document overlap. A model, not the annotator, then tabulated, per group, the selector,
exact prevalence, blinded label evidence, static-slice overlap and the auditor's stated
false positives, and proposed a verdict. The annotator read the tabulation and adopted
the proposed verdicts without deliberating each one independently, although the labels
themselves are human. Of the \adjndecisions{} decisions, \adjngrouprows{} cover whole groups and
\adjnmemberrows{} cover the members of \adjsplitgroup{}, which was decided per candidate
because its members' defect rates ranged from \adjsplitpmin{} to \adjsplitpmax. In total
there are \adjnapprove{} \texttt{approve}, \adjnreject{} \texttt{reject} and
\adjnunsure{} \texttt{unsure} decisions. With labels, the rule approves at least \thresholdaccept{} on at least
five labels and rejects below it. Without labels, it rejects a selector its own auditor
disqualified (a URL-keyword proxy, a closed list of domains from its own sample, a
match on two documents) and otherwise marks it \texttt{unsure}. The one exception is the
\texttt{slo.*} farm (G05), which at \valunsup{} is marked \texttt{unsure} rather than rejected so
that the unsupported finding stays visible. Two rows were corrected after adoption when
seven labels lost in our tabulation were restored and attributed by selector (Appendix~\ref{app:annotation});
the correction adds one validated selector to each of \textsc{Random} and
\textsc{LowConf} in RQ4 and makes \textsc{LowConf} the best condition per run.

\begin{table*}[t]
\centering\footnotesize
\providecommand{\ci}[1]{{\scriptsize\textcolor{rule!55!black}{[#1]}}}
\setlength{\tabcolsep}{3.4pt}\renewcommand{\arraystretch}{1.12}
\begin{tabular}{@{}llclrrrrl@{}}
\toprule
\textbf{Finding} & \textbf{Group} & \textbf{Runs} & \textbf{Baselines} &
\textbf{Prev.\,\%} & \textbf{$n$} & \textbf{$P$ [95\% CI]} & \textbf{Ov.\,\%} & \textbf{Status} \\
\midrule
\texttt{firebase\_content\_farm\_spam} & G01 & 3/3 & LD, LC, R & 5.32 & 19 & .74\,\ci{.51,.88} & 13 & \accepted \\
\rowcolor{zebra}\texttt{adult\_spam\_farm\_wordsalad} & G04 & 1/3 & -- & 0.39 & 19 & .84\,\ci{.62,.94} & 11 & \accepted \\
\rowcolor{zebra}\texttt{porn\_spam\_keyword\_farm} & G04 & 1/3 & -- & 0.36 & 12 & .83\,\ci{.55,.95} & 11 & \accepted \\
\texttt{slo\_prefix\_machine\_translated\_farm} & G05 & 1/3 & -- & 0.28 & 14 & .07\,\ci{.01,.31} & 0 & \candidate \\
\rowcolor{zebra}\texttt{porn\_doorway\_spam\_network} & G04 & 1/3 & -- & 0.20 & 20 & 1.00\,\ci{.84,1.00} & 10 & \accepted \\
\rowcolor{accsoft}\texttt{czech\_slovak\_code\_switched\_docs} & G02 & 1/3 & -- & 0.17 & 21 & .81\,\ci{.60,.92} & 100 & \accepted \\
\rowcolor{accsoft}\texttt{czech\_text\_mislabeled\_as\_slovak} & G02 & 1/3 & -- & 0.12 & 17 & .53\,\ci{.31,.74} & 100 & \accepted \\
\bottomrule

\end{tabular}
\caption{The seven distinct findings from three agent runs. \textbf{Group}:
phenomenon group assigned mechanically by document overlap (Jaccard $\geq 0.8$) or
nesting; rows sharing a group are one phenomenon, so the seven findings are four
defects. \textbf{Runs}: agent runs that produced the finding. \textbf{Baselines}:
baselines that proposed an equivalent slice (Jaccard $\geq 0.8$): R \textsc{Random},
LC \textsc{LowConf}, LD \textsc{LID-Disagree}, HD \textsc{HighDup}.
\textbf{Prev.}: exact prevalence over all \suitedocs{} documents. \textbf{$n$}:
blinded labels. \textbf{$P$}: human-judged defect rate with Wilson interval.
\textbf{Ov.}: largest share of the finding's documents that the \textsc{LowConf},
\textsc{LID-Disagree} or \textsc{HighDup} slice retrieves; the three G04 selectors lie
entirely inside the profanity slice $\texttt{n\_bad\_words} \geq 3$ (\S\ref{sec:results}).
\textbf{Status}: as released in the suite. The released suite also holds two metadata
invariants that are not slice findings (\S\ref{sec:results}).}
\label{tab:findings}
\end{table*}

\section{Results}
\label{sec:results}

\begin{figure}[t]
  \centering
  \includegraphics[width=\columnwidth]{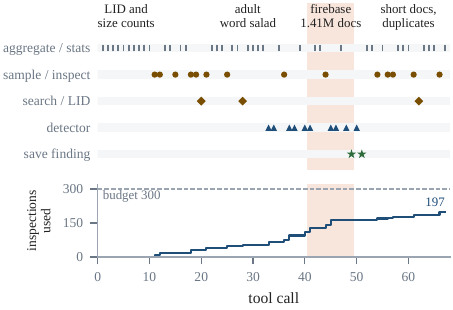}
  \caption{The 67 events logged for the main run (65 audit tool calls plus a \texttt{run\_detector} record per \texttt{save\_finding}), by tool type (top) and with cumulative document inspections against the budget of \budget{} (bottom). The shaded span, events 41 to 49, is the firebase investigation. The agent counts the \texttt{*.firebaseapp.com} and \texttt{*.web.app} hosts (\farmdocs{} documents), samples and evaluates the detector on them, and saves the finding at event 49. Run total 197 inspections.}
  \label{fig:trajectory}
\end{figure}

\begin{figure}[t]
  \centering
  \includegraphics[width=\columnwidth]{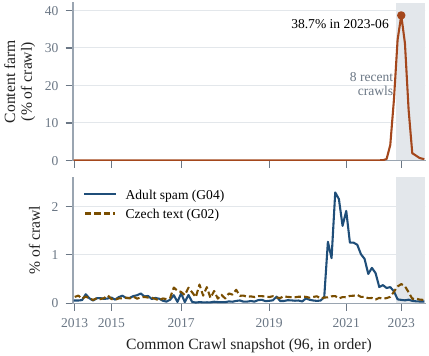}
  \caption{Share of each of the 96 Common Crawl snapshots matched by the accepted selectors, in crawl order. The Firebase content farm (top, \farmpct\% of the corpus overall) is almost absent before 2022 and reaches 38.7\% of the 2023-06 crawl. The adult spam family (union of the three G04 selectors) peaks around 2020 and falls from 0.53\% of older documents to 0.06\% of the eight most recent crawls (shaded), while Czech text (union of the G02 selectors) stays below 0.4\% throughout.}
  \label{fig:temporal}
\end{figure}

\paragraph{RQ1. The agent finds systematic defects that a speaker confirms.}
Across three runs the agent used 197, 180 and 173 of its \budget{} inspections and
recorded 2, 4 and 3 findings. Figure~\ref{fig:trajectory} follows run~1 from its first
free aggregate query to the saved content-farm finding at logged event 49, by which point it had
spent 162 inspections. Run~1 found the content farm and an adult keyword farm.
Run~2 found the farm, Czech text labelled Slovak, an adult doorway network and a
machine-translated farm on \texttt{slo.*} hosts. Run~3 found the farm, an adult
word-salad farm and Czech/Slovak code-switched documents. Deduplicating selectors leaves
seven, which collapse by document overlap and nesting into four phenomena
(Table~\ref{tab:findings}): content farms on free-hosting subdomains (G01), Czech or
code-switched documents labelled Slovak (G02), keyword-stuffed adult spam (G04) and the
\texttt{slo.*} farm (G05). The largest is a content farm on
\texttt{*.firebaseapp.com} and \texttt{*.web.app} with \farmdocs{} documents
(\farmpct\%) over 2{,}851 hosts. The farm
is nearly absent before 2022, rises sharply from late 2022, peaks at 38.7\% of the
2023-06 crawl and is still present in the last crawl, while the adult
spam peaks in 2020 and then fades (Figure~\ref{fig:temporal}). At 1{,}011 words on
average against a corpus mean of 438 it contributes \farmwordshare\% of all running
words. All three runs found it
independently, and we re-derived every statistic ourselves. The agent also volunteers
evidence against its own hypotheses (Appendix~\ref{app:counter}).

\begin{figure}[t]
  \centering
  \includegraphics[width=\columnwidth]{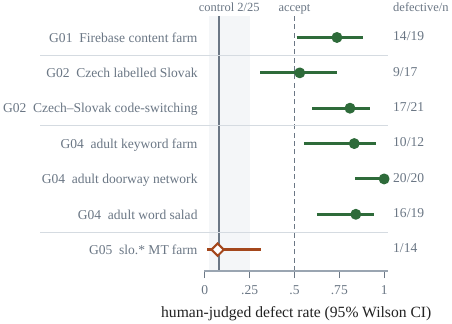}
  \caption{Human-judged defect rate for each of the seven agent findings, grouped by phenomenon. Bars are 95\% Wilson intervals and the right-hand column gives defective documents over documents judged. The shaded band is the interval for \valcontroln{} uniformly sampled documents (rate \valcontrol{}) and the dashed line is the acceptance threshold of \thresholdaccept{}. Six findings clear it. The \texttt{slo.*} selector (open marker, \valunsup{}) is indistinguishable from the random control.}
  \label{fig:validation}
\end{figure}

Blinded human labels (Figure~\ref{fig:validation}) support \validagent{} of the \nretained{} findings at
\valpmin--\valpmax{}, against \valcontrol{} in the \valcontroln{}-document control
(interval to \valcontrolhi{}), so \poolagentgroups{} of the four phenomena are supported. The labels
do not support the \texttt{slo.*} farm, which comes out at \valunsup{}, and it ships with
status \texttt{candidate} (\S\ref{sec:setup}).

\paragraph{RQ2. The compiled tests are cheap.}
All seven selectors are metadata-only, so the suite is exact over the full corpus.
It runs over all \suitedocs{} documents in \suitesecs\,s of wall-clock time on one
machine (about 13\,s of CPU time across 20 cores), without GPU
or LLM calls. Classifying every document with the same model instead costs
\$\clsperk{} per thousand documents, measured on 400-character snippets batched twenty
per call. That is \$\clsfull{} per pass, or \$\clstoken{} at raw token prices without
harness overhead, and it recurs every crawl. The comparison is not like for like. The
tests are cheap because they reuse columns the pipeline and the redistributor already
computed in corpus-wide passes (language identification, auxiliary identifiers,
profanity counts, MinHash clusters), and a selector over raw text would cost a text
scan. The LLM writes the test once, and that test runs at scale.

\begin{figure}[t]
\centering
\includegraphics[width=\columnwidth]{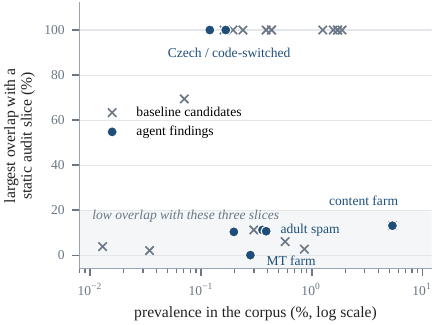}
\caption{Where the discovered slices sit relative to three static audit slices. Each
point is a candidate slice, placed by its prevalence in the corpus and by the largest
share of its documents that \textsc{LID-Disagree}, \textsc{LowConf} or
\textsc{HighDup} already retrieves. Agent findings are filled circles and baseline
candidates, the selectors the four static audits proposed at the same inspection
budget, are crosses. The two Czech findings are inside the language-ID disagreement
slice, as one would expect of a mislabelling defect. The content farm, the adult-spam
family and the machine-translation farm are not, although the adult-spam family lies
entirely inside a profanity-count slice not drawn here (\S\ref{sec:results}).}
\label{fig:overlap}
\end{figure}

\begin{figure}[t]
\centering
\includegraphics[width=\columnwidth]{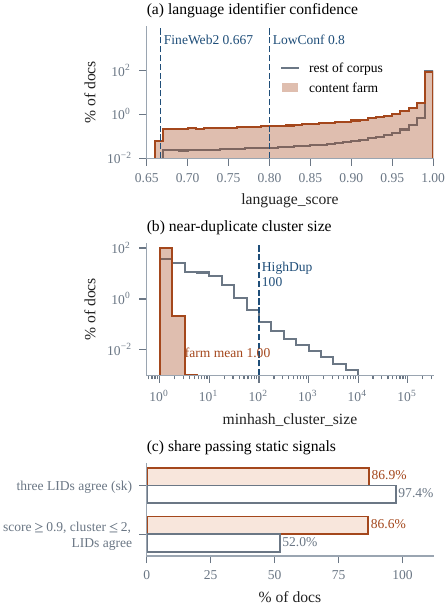}
\caption{The content farm (\farmdocs{} documents on \texttt{*.firebaseapp.com} and \texttt{*.web.app}) against the rest of the corpus on the three static signals. Panels (a) and (b) show the share of documents per bin on a log scale, with the \textsc{LowConf} cutoff of 0.8, the FineWeb2 threshold of 0.667 and the \textsc{HighDup} threshold of 100 dashed. No farm document reaches the duplication threshold. Panel (c) shows the share on which all three auxiliary identifiers return Slovak, and the share that also has \texttt{language\_score} of at least 0.9 and a cluster size of at most 2, which is \farmpassall\% of the farm (\farmpassalldocs{} documents).}
\label{fig:staticmiss}
\end{figure}

\paragraph{RQ3. Static strategies reach every validated phenomenon.}
The question here is which \emph{static audit} would have led to the seven findings.
The two Czech findings sit entirely inside
the \textsc{LID-Disagree} slice (overlap 100\%), as the agent built them from the
auxiliary LID columns, the failure mode a disagreement audit is designed to catch. The
adult-spam family (G04) overlaps the three slices of \S\ref{sec:setup} by at most
11\%, but its selectors threshold the corpus's own profanity count: the slice
$\texttt{n\_bad\_words} \geq 3$ (\profslicepct\% of the corpus) contains every G04
selector, and the \profbudget{} most profane documents of the text subsample belong to
G04 and to no other agent finding (Appendix~\ref{app:annotation}). We did not run the
model on this profanity-ranked sample, so whether it would have proposed and measured
the slice is untested. The \texttt{slo.*} farm falls outside every slice,
with only 12 disagreements in 73{,}835 documents (Figure~\ref{fig:overlap}), but
the labels do not support it.

The content farm is the one case where static \emph{filtering} fails
(Figure~\ref{fig:staticmiss}). Its mean \texttt{language\_score} is 0.98, well clear of
the 0.8 cutoff, its mean MinHash cluster size is 1.00, so \emph{no} farm document
reaches the duplication threshold, and all three auxiliary identifiers agree with the
label on 87\% of them. Even at stricter thresholds (a score of at least 0.9, a cluster
of at most two, all three identifiers agreeing), \farmpassalldocs{} of its \farmdocs{}
documents (\farmpassall\%) pass, as do \restpassall\% of the remaining documents, so
these signals cannot single the farm out from the rest of the corpus. Static \emph{sampling}
does reach it. At \farmpct\% prevalence a \budget{}-document draw holds ${\sim}16$ farm
documents, and three baselines proposed it (RQ4). The answer to RQ3 is therefore no:
each validated phenomenon is retrievable from one simple shipped signal (LID disagreement for G02, the profanity
count for G04, uniform sampling for G01), and the agent
contributed the measurement and characterisation.

A fourth run with \texttt{claude-opus-5} (Appendix~\ref{app:robust}) reproduced the
content-farm slice exactly and reported a metadata invariant: 10{,}691{,}128 documents
(\invalidshare\%) carry a \texttt{language\_score} above 1.0, up to 1.00001. This is a
numerical convention rather than contamination (Appendix~\ref{app:crosslang}), yet it
matters to anyone who treats the field as a probability, for instance when
re-thresholding with FineWeb2's per-language cutoffs. We release it as an invariant
test instead of a filter. The suite holds one more invariant, written by us with the
framework rather than found by an agent, that document identifiers are unique. Thirteen
identifiers occur twice in the release (26 rows), all in CC-MAIN-2024-10, the
second-newest crawl
(Appendix~\ref{app:heldout}).

\begin{table}[t]
\centering\footnotesize
\providecommand{\grouphead}[1]{{\scriptsize\sffamily\bfseries\textcolor{acc}{#1}}}
\setlength{\tabcolsep}{2.4pt}\renewcommand{\arraystretch}{0.92}
\begin{tabular}{@{}lrrrr@{\hspace{7pt}}lr@{}}
\toprule
\textbf{Condition} & \textbf{Insp.} & \textbf{Cand.} & \textbf{Val.} & \textbf{Grp.} &
\textbf{G\#} & \textbf{USD} \\
\midrule
\rlap{\grouphead{Baselines (single runs)}} & & & & & & \\
\textsc{Random} & 300 & 7 & 2 & 2 & 1,2 & 1.22 \\
\textsc{LowConf} & 300 & 6 & 4 & 3 & 1,2,9 & 1.03 \\
\textsc{LID-Disagree} & 300 & 5 & 2 & 2 & 1,2 & 1.16 \\
\textsc{HighDup} & 300 & 5 & 2 & 2 & 2,10 & 1.06 \\
\midrule
\rlap{\grouphead{Agent (single runs)}} & & & & & & \\
\rowcolor{accsoft}Agent run 1 & 197 & 2 & 2 & 2 & 1,4 & 2.21 \\
\rowcolor{accsoft}Agent run 2 & 180 & 4 & 3 & 3 & 1,2,4 & 2.84 \\
\rowcolor{accsoft}Agent run 3 & 173 & 3 & 3 & 3 & 1,2,4 & 2.45 \\
\midrule
\rlap{\grouphead{Pooled}} & & & & & & \\
\rowcolor{zebra}All 4 baselines & 1{,}200 & 21 & 6 & 4 & 1,2,9,10 & 4.46 \\
\rowcolor{zebra}All 3 agent runs & 550 & 7 & 6 & 3 & 1,2,4 & 7.50 \\
\bottomrule

\end{tabular}
\caption{Budget-matched comparison. Rows above the pooled block are single runs, each
reading at most \budget{} documents; pooled rows aggregate the four baselines and the
three agent runs, counting selectors that match the same documents within one group
once. \textbf{Insp.}: documents read. \textbf{Cand.}: distinct candidate selectors.
\textbf{Val.}: selectors with an adopted \texttt{approve} verdict on
five or more blinded labels at a defect rate of at least \thresholdaccept{}
(\S\ref{sec:setup}); a selector two conditions proposed identically is credited to
both. \textbf{Grp.}, \textbf{G\#}: phenomenon groups as numbered in
Table~\ref{tab:findings} (1 = G01, \dots; G09 is an adult tag farm, G10 an injected
advertising widget). \textbf{USD}: LLM cost.}
\label{tab:budget}
\end{table}

\begin{figure}[t]
  \centering
  \includegraphics[width=\columnwidth]{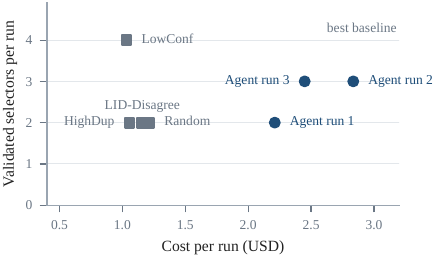}
  \caption{Validated selectors per run against the cost of the run. Agent runs are blue circles and static baselines grey squares. The best baseline, \textsc{LowConf}, validates four selectors over three phenomenon groups at \$\baseusdmin{}, more than any agent run at over twice the cost.}
  \label{fig:perrun}
\end{figure}

\paragraph{RQ4. Per run, the best baseline validates more than the agent.}
At \budget{} inspections each, the static strategies proposed \basecandsum{} selectors
counted per baseline, \poolbasecand{} of them distinct
(two identical to an agent selector). Table~\ref{tab:budget}
compares every run with every baseline.
Each agent run validated \runvalidmin--\runvalidmax{} selectors over
\rungroupsmin--\rungroupsmax{} groups, while the baselines validated
\basevalidmin--\basevalidbest{}. The best, \basevalidbestname{}, validates more
selectors than any agent run (\basevalidbest{} against \runvalidmax{}) over as many
groups (three) at under half the cost of each (Figure~\ref{fig:perrun}).
\textsc{Random} is credited with the farm and with an English-text selector because
both are byte-identical to selectors another condition proposed, and
\gtwobaselines{} baselines produced their own validated selectors in the
mislabelled-language group (G02) at \gtwobasepmin--\gtwobasepmax{}. Pooled, the four
baselines cover \poolbasegroups{} groups with \poolbasevalid{} validated selectors, and
the three agent runs \poolagentgroups{} groups with \poolagentvalid{}. The groups differ. Only the agent validated
\agentonlygroups{} (\gfourpmin--\gfourpmax\% of the corpus per selector), but
\textsc{LID-Disagree} proposed a selector in the same group (\texttt{\ldadultsel},
\ldadultpct\%) that was approved with its group G04 although it drew no labels of its
own, so it does not count, and the group is inside the profanity slice (RQ3). Only the
baselines validated \baseonlygroups{}, an adult tag farm of \gninedocs{} documents
(\gnineprec{}, $n=\gninen$) and an advertising widget injected across unrelated hosts.

What differs is what comes with each selector. Both kinds carry an expected precision and
counter-evidence, but the baselines judged theirs from 400-character views of
\budget{} documents, while every agent selector also arrives with an exact prevalence
measured on the whole corpus. That
characterisation is work a curator would otherwise do by hand, and an agent run costs
\$\runusdmin--\$\runusdmax{} against \$\baseusdmin--\$\baseusdmax{} per baseline.

\paragraph{Held-out split.}
The selectors were written from the text subsample of the training split, so they
could be fitted to the shards the agent saw. All \suiten{} tests run
unchanged on an independent held-out split of 28{,}720 documents that the release
defines and rebuilds and the agent never touched (Table~\ref{tab:heldout}). This is
evidence against overfitting to the explored shards, not of stability over time
(Appendix~\ref{app:temporal}).

\begin{table}[t]
\centering\footnotesize
\setlength{\tabcolsep}{4pt}
\begin{tabular}{@{}lrrr@{}}
\toprule
\textbf{Test} & \textbf{train} & \textbf{held-out} & \textbf{ratio} \\
 & \textbf{\%} & \textbf{\%} & \\
\midrule
{\scriptsize\ttfamily language\_score\_not\_a\_probability} & 40.389 & 40.230 & 1.00 \\
\rowcolor{zebra}{\scriptsize\ttfamily firebase\_content\_farm\_spam} & 5.321 & 5.024 & 0.94 \\
{\scriptsize\ttfamily adult\_spam\_farm\_wordsalad} & 0.387 & 0.348 & 0.90 \\
\rowcolor{zebra}{\scriptsize\ttfamily porn\_spam\_keyword\_farm} & 0.359 & 0.334 & 0.93 \\
{\scriptsize\ttfamily slo\_prefix\_machine\_translated\_farm} & 0.279 & 0.292 & 1.05 \\
\rowcolor{zebra}{\scriptsize\ttfamily porn\_doorway\_spam\_network} & 0.198 & 0.171 & 0.86 \\
{\scriptsize\ttfamily czech\_slovak\_code\_switched\_docs} & 0.167 & 0.178 & 1.06 \\
\rowcolor{zebra}{\scriptsize\ttfamily czech\_text\_mislabeled\_as\_slovak} & 0.120 & 0.118 & 0.99 \\
{\scriptsize\ttfamily document\_ids\_unique} & 0.000 & 0.000 & -- \\
\bottomrule

\end{tabular}
\caption{The released suite on the held-out split. Prevalences are exact counts on
each split. Every slice prevalence lands within
0.86--1.06 times its training value; the invalid-confidence invariant fails there too,
while the identifier invariant passes, as expected of 13 collisions in \corpusM{}M
training documents (Appendix~\ref{app:heldout}).}
\label{tab:heldout}
\end{table}

\section{Conclusion}
\label{sec:conclusion}
An LLM agent audited Slovak FineWeb2 for \$\runusdmin--\$\runusdmax{} a run. It found,
measured and characterised slices that a native speaker then confirmed, and left a test
suite that runs without a model. It validated fewer selectors per run than the same
model reading a well-chosen static sample, and every defect it validated was within
reach of one static strategy (we did not run the model on the profanity-ranked sample
that reaches the adult spam, \S\ref{sec:results}). The contribution is the
workflow, which turns a one-off audit into executable, validated tests and can fail a
finding, as it did for one of ours (shipped as a candidate). It already runs
on eleven other FineWeb2 configurations (Appendix~\ref{app:crosslang}), and another community can point the adapter at its
own corpus and run the suite.

\section*{Limitations}

\paragraph{Scope.} This study covers one language, one corpus and three runs of one
model. Agentic discovery is shown only on Slovak, and the other languages in
Appendix~\ref{app:crosslang} only re-ran tests written for Slovak. A fourth,
stronger-model run reproduced the largest phenomena outside the frozen protocol, but
its extra findings are unvalidated. The agent used 58--66\% of its allowance.

\paragraph{Validation.} Human validation covers \vallabels{} documents labelled by one
native speaker (\valcellmin--\valcellmax{} per finding, median \valcellmed), which
limits precision estimates and gives wide intervals.\iaasentence{}\codingrulesentence{}
The single-coder design does not measure agreement on ambiguous Czech/Slovak text.
Among the 25 documents drawn per agent finding, at least \annminhosts{} hosts are
represented and no host contributes more than \anntophost\%.
Validation measures precision only. We do not know the recall of any selector, so the
tests guard specific patterns and do not show that they capture contamination
broadly. Our defect criterion (not Slovak or junk) is coarse, and we do not measure
downstream language-model effects \citep{artetxe-etal-2022-corpus}.

\paragraph{Comparison.} As \S\ref{sec:setup} notes, the baselines lack the agent's tools as well as its adaptivity. A
profanity-ranked baseline was not run with the model, and we report only the slice
analysis, which shows that the G04 selectors threshold \texttt{n\_bad\_words}, a column
the release already ships. Some baseline candidates drew no labels, so their status
rests on the no-labels branch of the adjudication rule, and every verdict was proposed
by a model and adopted rather than deliberated independently. Two of those verdicts
were later corrected by the stated rule, at the author's direction and without a new
deliberation, after labels lost in our tabulation were restored,
and the correction moved the best baseline from tying the agent to exceeding it.

\paragraph{Cost and stability.} The cost comparison relies on metadata the pipeline
already computed (\S\ref{sec:results}). All retained
selectors use metadata only, so their prevalences are exact, whereas a text-dependent selector
would use the text subsample and require extrapolation. The prevalence guards
are calibrated for corpus-wide evaluation, and on a single crawl natural variation trips
all seven guards and can produce false alarms. The detectors are cheap review aids that
do not guarantee stability, and some host patterns will drift as the web changes.

\section*{Ethical Considerations}

The corpus is public web text redistributed under ODC-By, and we report aggregate
statistics and short excerpts only. Several discovered slices consist of adult
spam. The corpus release already flags profanity-list matches, and we neither
reproduce such content in the paper nor include it in the released examples
beyond what is necessary to document the finding. The annotation task required a
native speaker to read web text that includes such material. Before the task began,
the annotator was told to expect explicit content and that any document could be
skipped and the work stopped at any time.
Released excerpts and trajectories are redacted for third-party personal
data. Real email addresses are replaced with a marker at the tool boundary, while
the corpus's own anonymisation placeholders are kept, since one candidate finding
concerns them. The placeholders appear in 3.74\% of sampled documents against
0.009\% carrying an address that survived the pipeline's scrubbing, so
FineWeb2's anonymisation is close to complete and we do not republish the residue.

\section*{Acknowledgments}

This work was partially supported by the European Union under the project lorAI -- Low Resource Artificial Intelligence, GA No.~\href{https://doi.org/10.3030/101136646}{101136646}; by ALFIE -- Assessment of Learning technologies and Frameworks for Intelligence, a project funded by European Union under the Horizon Europe, GA No.~\href{https://doi.org/10.3030/101177912}{101177912}; and carried out in the framework of CLARIN-SK infrastructure, the national node of the CLARIN-ERIC, and was supported by the Ministry of Education, Research, Development, and Youth of the Slovak Republic.

\bibliography{custom}

\appendix

\section{Audit Tool API}
\label{app:tools}
The eleven tools exposed to the agent and their cost against the inspection
budget, under which aggregation is free and reading documents is not.

{\footnotesize
\providecommand{\costfree}{\statustag{run}{runsoft}{free}}
\providecommand{\costdoc}{\statustag{acc}{accsoft}{1 inspection per document}}
\begin{description}[leftmargin=0.8em, labelindent=0pt, itemsep=2pt, parsep=0pt, topsep=2pt, style=nextline, font=\normalfont]
\item[{\parbox[t]{\linewidth}{\raggedright\textcolor{acc}{\texttt{schema()}}}}] \costfree\enspace Describe the available relations, columns, and the remaining inspection budget.
\item[{\parbox[t]{\linewidth}{\raggedright\textcolor{acc}{\texttt{aggregate(sql)}}}}] \costfree\enspace Run a read-only SQL query over the corpus relations (counts, distributions, group-bys).
\item[{\parbox[t]{\linewidth}{\raggedright\textcolor{acc}{\texttt{sample\_documents(condition, n, seed)}}}}] \costdoc\enspace Draw a random sample of documents matching a SQL condition, returning metadata and a 400-character snippet each.
\item[{\parbox[t]{\linewidth}{\raggedright\textcolor{acc}{\texttt{inspect\_document(doc\_id)}}}}] \costdoc\enspace Read one document in full (up to 6000 characters) together with all its metadata.
\item[{\parbox[t]{\linewidth}{\raggedright\textcolor{acc}{\texttt{search\_text(pattern, condition, n)}}}}] \costdoc\enspace Count and retrieve documents whose text matches a regular expression.
\item[{\parbox[t]{\linewidth}{\raggedright\textcolor{acc}{\texttt{domain\_stats(condition, n)}}}}] \costfree\enspace Most frequent hosts among documents matching a condition, with each host's share of its own documents.
\item[{\parbox[t]{\linewidth}{\raggedright\textcolor{acc}{\texttt{crawl\_stats(condition)}}}}] \costfree\enspace Rate of documents matching a condition in each crawl snapshot.
\item[{\parbox[t]{\linewidth}{\raggedright\textcolor{acc}{\texttt{sentence\_lid(doc\_id)}}}}] \costdoc\enspace Run a sentence-level language identifier over one document and report the language of each line plus the aggregate distribution.
\item[{\parbox[t]{\linewidth}{\raggedright\textcolor{acc}{\texttt{run\_detector(name, selector)}}}}] \costfree\enspace Measure how much of the corpus a candidate rule (SQL condition) selects, exactly on the full corpus when the rule uses metadata only.
\item[{\parbox[t]{\linewidth}{\raggedright\textcolor{acc}{\texttt{evaluate\_detector(name, selector, n, seed)}}}}] \costdoc\enspace Inspect a random sample of the documents a candidate rule selects, to check its precision.
\item[{\parbox[t]{\linewidth}{\raggedright\textcolor{acc}{\texttt{save\_finding(finding\_id, description, selector, evidence, example\_ids, expected\_precision, likely\_false\_positives, assertion)}}}}] \costfree\enspace Record a supported finding as a machine-readable corpus test.
\end{description}

}

\section{Agent Prompt}
\label{app:prompt}
The agent's complete system prompt, verbatim:
\begin{artefactbox}{System prompt (verbatim)}
{\footnotesize\begin{verbatim}
You are a careful empirical researcher
auditing a large text corpus that was built
automatically from web crawls. You interact
with the corpus only through the provided
tools. You cannot see the corpus any other
way, and you cannot modify it.

The corpus under audit is one language-
specific slice of a large multilingual web
corpus. Its curation pipeline labelled every
document in it as Slovak (`slk_Latn`),
applied quality heuristics, and deduplicated
it. The slice is intended to be used as
pretraining data for language models and as
a linguistic resource for Slovak.

Your job is to find out, empirically, in
what ways this slice fails to be what it
claims to be, and to leave behind machine-
checkable evidence of each failure you can
substantiate.

# What counts as a finding

A finding is a *systematic* property of a
*coherent subset* of documents that makes
those documents wrong, misleading, or
unusable given what the corpus claims to be.
It must be:

1. **Systematic** -- it recurs across
   documents, hosts, or crawls; a single
   strange document is not a finding.
2. **Reproducibly retrievable** -- you can
   write a SQL condition over the available
   columns (optionally using the text) that
   selects the affected documents with
   reasonable precision.
3. **Grounded** -- you have actually
   inspected documents that the condition
   selects, and you know what the condition
   wrongly catches.
4. **Quantified** -- you have measured how
   much of the corpus it affects.

Do not invent categories a priori and go
looking for them. Start from what the data
actually shows: distributions, outliers,
structure across hosts and crawls, and
documents you have read. Let the corpus
tell you what is wrong with it.

# How to work

Iterate, and be explicit at each step:

1. Orient: look at the schema and at
   aggregate structure.
2. Form a specific hypothesis about a
   defect, stated so that it could be
   wrong.
3. Retrieve targeted evidence *and*
   counter-evidence. Look at documents
   that should be affected and documents
   that should not.
4. Refine or reject the hypothesis.
   Rejecting a hypothesis on evidence is a
   good outcome; say so and move on.
5. Estimate prevalence with
   `run_detector`.
6. Where the phenomenon admits it,
   synthesise a cheap selector, check its
   precision yourself with
   `evaluate_detector`, and record it with
   `save_finding`, including what it
   wrongly catches.

# Constraints

* You have a strict budget of document
  inspections, shown by `schema` and after
  each inspecting call. Spend it
  deliberately: aggregates are free,
  reading documents is not. Reserve
  budget for checking your candidate
  selectors, not only for exploring.
* Aim for a small number of strong,
  distinct findings rather than many weak
  ones. Do not split one phenomenon into
  arbitrary variants; one finding per
  phenomenon, with the best selector you
  can write.
* A selector that only uses metadata
  columns can be measured exactly on the
  whole corpus, which is more useful than
  one that needs the text. Prefer cheap
  selectors when they are about as
  accurate.
* Be calibrated. When you record a
  finding, your `expected_precision`
  should be your honest estimate from the
  documents you inspected, and your
  `evidence` field should state what you
  checked, including anything that argues
  against the finding.
* Never claim a measurement you did not
  make.

Work autonomously until your inspection
budget is nearly exhausted or you have no
further productive hypotheses. Then stop
and summarise: the findings you recorded,
the hypotheses you rejected and why, and
what you would investigate next with more
budget.
\end{verbatim}}
\egroup 
\end{artefactbox}

\section{What the Agent Was Told}
\label{app:integrity}
Two claims carry the experiment, so a script (\texttt{check\_experiment\_integrity.py})
checks both.

\paragraph{No taxonomy leakage.} The prompt is tested against
41 terms naming defects, failure modes, or the
neighbouring language, including every phenomenon the agent ended up
reporting, and contains none of them, nor any instruction of the form "look for"
or "typical defects". It does state that the slice claims to be
Slovak, without which nothing could be judged, and what the pipeline did
(language identification, quality heuristics, deduplication). The column
documentation likewise explains the corpus's own published fields, including that
\texttt{minhash\_cluster\_size} counts near-duplicates and
\texttt{n\_bad\_words} counts profanity-list matches, which is a data
dictionary given identically to the baselines. Describing what a shipped column
counts does not tell the agent that duplication or profanity is a defect, although
it does make the profanity count visible, and the adult-spam selectors use it
(\S\ref{sec:results}).

\paragraph{No tools beyond the audit API.} Scanning the recorded streams of the
three frozen runs shows that every one of the 227 tool calls is an
\texttt{mcp\_\_audit\_\_*} call, except one call per run to the harness's own
tool-selection routine, which takes a list of tool names and returns their
schemas. No shell, file access, network, or built-in retrieval tool was invoked.

\section{Why the Czech Selector Needs Three Identifiers}
\label{app:chalis}
The Czech findings rest on three auxiliary identifiers agreeing, so the quality of
that rule needs its own check. CHALIS \citep{tichy2026chalis} is a benchmark of
deliberately hard language-identification cases that separates Czech/Slovak sentences
with a single correct label from sentences that are equally valid in both
languages, so it can calibrate the rule without asking a person.

No single identifier is defensible here. On the 36 single-label Slovak
sentences, fastText is right 22.2\% of the time, langdetect
69.4\% and Lingua 66.7\%; the majority
class is Czech, and the Slovak side is where they fail. The conjunction is different. Requiring all three to say Czech fires on 257 of the
527 single-label sentences and is correct on
256 of them, for 99.6\%
precision on an external benchmark built to break language identification.

Two caveats keep this from standing in for human validation. CHALIS scores
sentences (median 24 characters) while our selector
scores documents, and our defect criterion is broader than language. And
55.4\% of the benchmark's Czech/Slovak
material is inherently ambiguous, and on those 655 jointly valid sentences the
conjunction still fires 28.5\% of the time. This is the
false-positive class the finding declares, short colloquial Slovak that reads as Czech
to every identifier, and it is why the precision of these two slices must be judged by a speaker.

\section{The Written Tests on Other Languages}
\label{app:crosslang}
This appendix tests whether the written tests and the framework carry over to other
languages, not whether the agent discovers defects there, as we ran no agent on them.
Nothing in the framework is Slovak except the tests, so we pointed it at the
{\tt test} shards of eleven other FineWeb2 configurations, plus FineWeb2's own
Slovak (our corpus is a redistribution of it with extra columns). Upstream ships
fewer columns, lacking the auxiliary LID predictions and the profanity count, which
is the situation a second language community is actually in. The adapter
materialises absent columns as NULL and says so, and the runner then refuses to
score a test it cannot measure. On upstream Slovak, five of the seven slice guards
are therefore reported unmeasurable instead of passing at zero prevalence, which
would have been the dangerous outcome. The two that survive on metadata alone
reproduce there, the content farm at 3.72\% and the machine-translation farm at
0.258\%, so those slices come from FineWeb2, not the redistributor. On the
11 foreign corpora all seven Slovak tests are correctly reported out of
scope, and the two language-neutral invariants run. Identifier uniqueness passes
everywhere, and the invalid-confidence invariant fails in 11 of 12 configurations
(Table~\ref{tab:crosslang}), with Swahili the one clean configuration at a
maximum score of 0.99990. The shares vary by a factor of thirty-five, but the
ceiling barely moves, and it has a mechanical cause.
fastText avoids $\log 0$ by computing $\log(p+10^{-5})$, so a maximally confident
label is reported as $p+10^{-5}$. The maximum we observe equals
\texttt{float32}$(1.00001)$ in 8 of the 11 affected configurations and lies within
$2\times10^{-6}$ of it in the other three (Amharic, Welsh and Hebrew). The pipeline's language filter stores that number
unchanged, which makes the defect an upstream convention faithfully passed through
rather than a mistake in any one language. Although it leaves every document intact,
it matters to a consumer who reads the field as a probability, for instance to
re-threshold with FineWeb2's per-language cutoffs or to average it as a confidence.
The fix is one line: clamp the score, or subtract the epsilon before publishing it.

\begin{table}[h]
\centering\footnotesize
\setlength{\tabcolsep}{3pt}
\begin{tabular}{@{}llrrrr@{}}
\toprule
\textbf{Config} & \textbf{Script} & \textbf{docs} & \textbf{$>$1.0} &
\textbf{\%} & \textbf{max} \\
\midrule
{\scriptsize\ttfamily mlt\_Latn} & Latin & 5{,}193 & 4{,}106 & 79.1 & 1.00001 \\
\rowcolor{zebra}{\scriptsize\ttfamily hye\_Armn} & Armenian & 16{,}131 & 10{,}281 & 63.7 & 1.00001 \\
{\scriptsize\ttfamily ces\_Latn} & Latin & 28{,}051 & 16{,}798 & 59.9 & 1.00001 \\
\rowcolor{zebra}{\scriptsize\ttfamily cym\_Latn} & Latin & 8{,}395 & 4{,}348 & 51.8 & 1.00001 \\
{\scriptsize\ttfamily khm\_Khmr} & Khmer & 16{,}337 & 8{,}369 & 51.2 & 1.00001 \\
\rowcolor{zebra}{\scriptsize\ttfamily isl\_Latn} & Latin & 32{,}586 & 15{,}215 & 46.7 & 1.00001 \\
{\scriptsize\ttfamily slk\_Latn} & Latin & 32{,}557 & 12{,}288 & 37.7 & 1.00001 \\
\rowcolor{zebra}{\scriptsize\ttfamily tha\_Thai} & Thai & 24{,}037 & 6{,}575 & 27.4 & 1.00001 \\
{\scriptsize\ttfamily nno\_Latn} & Latin & 12{,}227 & 751 & 6.1 & 1.00001 \\
\rowcolor{zebra}{\scriptsize\ttfamily amh\_Ethi} & Ethiopic & 3{,}973 & 152 & 3.8 & 1.00001 \\
{\scriptsize\ttfamily heb\_Hebr} & Hebrew & 26{,}089 & 597 & 2.3 & 1.00001 \\
\rowcolor{zebra}{\scriptsize\ttfamily swh\_Latn} & Latin & 12{,}669 & 0 & 0.0 & 0.99990 \\
\bottomrule

\end{tabular}
\caption{\texttt{language\_score} above 1.0 in upstream FineWeb2, measured on
each configuration's {\tt test} shard. \textbf{max} is the largest score
observed. No shard contains a duplicate identifier.}
\label{tab:crosslang}
\end{table}

\section{More on the Held-out Split}
\label{app:heldout}
The invalid-confidence invariant fails on the held-out data too, at 40.23\%. The
identifier-uniqueness invariant passes there, as one would expect of 13 collisions in
\corpusM{}M documents. All of them sit in CC-MAIN-2024-10, the second-newest crawl, and in every case both
rows come from the same WARC file with different extracted text, so a join on
\texttt{id} fans out instead of deduplicating. Building the adapter over the held-out
split takes seconds, which is also the shortest demonstration that a second corpus
implementing the adapter interface needs no change to the suite.

\section{Where the Slices Live in Time}
\label{app:temporal}
A guard is only useful on a future crawl if the defect is still there, so we ran
every test on each of the 96 snapshots separately (Table~\ref{tab:temporal}).
Five of the seven slice guards appear in at least half the crawls, and the two
Czech slices are present in every one of the 96 (at roughly 0.06--0.40\% per crawl), so they are persistent properties of the
pipeline and not incidents. The content farm is the opposite and more interesting.
It appears in only 23 crawls but is 21.6 times larger in the eight
newest than in everything before (15.9\% against 0.74\%). It rose sharply from late
2022, peaked at 38.7\% of the 2023-06 snapshot and was still present in the last
crawl at 0.39\%, which is exactly
when a regression test earns its keep. The adult-spam family runs the other way,
peaking in 2020 and nearly gone by the end, so those guards document something
that was, instead of filtering something that matters now. The corpus-wide prevalence
alone would not have told us which was which.

\begin{table*}[t]
\centering\footnotesize
\setlength{\tabcolsep}{5pt}
\begin{tabular}{@{}lrrrrrr@{}}
\toprule
\textbf{Test} & \textbf{all} & \textbf{old} & \textbf{new} & \textbf{ratio} & \textbf{crawls} & \textbf{peak} \\
\midrule
{\scriptsize\ttfamily language\_score\_not\_a\_probability} & 40.39 & 40.50 & 40.13 & 0.99$\times$ & 96/96 & 2024-10 (46.9) \\
\rowcolor{zebra}{\scriptsize\ttfamily firebase\_content\_farm\_spam} & 5.32 & 0.74 & 15.91 & \cellcolor{warnsoft}21.6$\times$ & 23/96 & 2023-06 (38.7) \\
{\scriptsize\ttfamily adult\_spam\_farm\_wordsalad} & 0.39 & 0.53 & 0.06 & \cellcolor{runsoft}0.11$\times$ & 96/96 & 2020-40 (2.3) \\
\rowcolor{zebra}{\scriptsize\ttfamily porn\_spam\_keyword\_farm} & 0.36 & 0.50 & 0.04 & \cellcolor{runsoft}0.07$\times$ & 95/96 & 2020-40 (2.3) \\
{\scriptsize\ttfamily slo\_prefix\_machine\_translated\_farm} & 0.28 & 0.27 & 0.30 & 1.1$\times$ & 46/96 & 2021-49 (0.9) \\
\rowcolor{zebra}{\scriptsize\ttfamily porn\_doorway\_spam\_network} & 0.20 & 0.28 & 0.01 & \cellcolor{runsoft}0.05$\times$ & 84/96 & 2020-40 (1.4) \\
{\scriptsize\ttfamily czech\_slovak\_code\_switched\_docs} & 0.17 & 0.15 & 0.22 & \cellcolor{warnsoft}1.5$\times$ & 96/96 & 2023-06 (0.4) \\
\rowcolor{zebra}{\scriptsize\ttfamily czech\_text\_mislabeled\_as\_slovak} & 0.12 & 0.14 & 0.08 & 0.56$\times$ & 96/96 & 2017-26 (0.4) \\
\bottomrule

\end{tabular}
\caption{Every test evaluated per crawl. \textbf{all}/\textbf{old}/\textbf{new}:
prevalence (\%) over the whole corpus, over all but the eight newest snapshots,
and over those eight. \textbf{ratio}: new over old. \textbf{crawls}: snapshots in
which the selector matches anything. \textbf{peak}: the snapshot where the slice
is largest, with its prevalence there (\%).}
\label{tab:temporal}
\end{table*}

\section{Findings as Patches}
\label{app:patches}
A test states that a slice should not be there, and the same selector used as a
filter is a patch. Applying all \nretained{} retained selectors, including the
\texttt{slo.*} candidate that the labels do not support, gives an upper bound on what
the audit would remove, dropping 1{,}615{,}885 of \suitedocs{} documents (6.10\%) and 16.31\% of
the words. The two figures differ because the slices run long, and the
\texttt{slo.*} machine-translation farm is 0.28\% of documents but 3.22\% of the
words. The release writes a drop list from the suite (\texttt{apply\_patches.py}), a
Parquet file of the matched identifiers, so a correction travels without the corpus.
Findings whose defect
lives in the metadata instead of the documents (the identifier collisions
and the out-of-range confidence) ship as tests only and never as filters, because deleting those rows
would hide the bug instead of reporting it.

We check rather than assume that the two halves agree, materialising the
patched corpus and re-running the whole suite over it. Every patched guard goes to
exactly zero matches, 93.90\% of documents survive, and the two non-filterable
invariants still fail, which is the intended asymmetry. As a side effect,
removing the slices raises the invalid-confidence share from \invalidshare\% to
41.5\%, because the spam the filter removes scores below the corpus average. That
defect is independent of the content defects, and only the pipeline, not a
filter, can fix it.

\section{Trajectory Replay}
\label{app:replay}
The trajectory log should make an investigation reproducible, so we
re-executed every deterministic read-only call
from the three frozen runs against the same corpus and compared the results with
what was logged. Of the \replayn{} logged calls, \replayset{} (\replaysetpct\%) return
the same rows as a set and \replayexact{} (\replayexactpct\%) are identical byte for
byte, while the remaining \replayorderonly{} set matches return the same rows in a
different order. The \replaysetdiff{} calls that differ as sets are all \texttt{LIMIT}
queries whose ordering key is tied at the
cutoff: two host-statistics calls where domains share a document count and one
agent-written query where six URLs tie, so SQL is free to return a different
tied subset. The host-statistics tool now imposes a total order, but we cannot
constrain queries the agent writes itself.

\section{Counter-evidence the Agent Volunteered}
\label{app:counter}
The agent volunteers evidence against its own hypotheses. In a pilot run it
reported that diacritic-free informal Slovak fools all three auxiliary
identifiers into voting Czech, concluded that roughly half of a strict
triple-agreement slice was therefore not a defect, and lowered its own expected
precision to 0.5 accordingly, a caution about LID-disagreement audits that we
never prompted for.

\section{Cross-model Robustness}
\label{app:robust}
A fourth run used the same environment, prompt and \budget{}-inspection budget with
\texttt{claude-opus-5} (\opusturns{} turns, \$\opususd). It is reported
separately because it is outside the frozen three-run protocol and its findings
were not part of the blinded annotation package. It recorded eight findings.
Measured by document-set overlap against the seven primary findings, the
firebase/web.app content farm is reproduced exactly (Jaccard 1.00, the
identical \farmdocs{} documents), the machine-translated health farm is found with a
broader selector (147{,}318 documents, Jaccard 0.31 with the
\texttt{slo.*} slice), and Czech mislabelling with a different selector
(23{,}076 documents, Jaccard 0.09). Four further candidate phenomena are new and
remain unvalidated: machine-translated \texttt{sk.}-locale subdomains of
foreign sites (903{,}273 documents, excluding \texttt{sk.wikipedia.org}),
flat-HTML offshore content farms (473{,}571), documents whose e-mail addresses
were silently replaced by placeholders (76{,}133), and advertising boilerplate
spliced into article text (18{,}887). The eighth is the metadata invariant
discussed in \S\ref{sec:results}, which we verified arithmetically and released
as a test. We draw one conclusion only: the largest phenomenon is a property of
the corpus and not of one model, and a stronger model proposes more candidates, at
\opusratiomin--\opusratiomax{} times the cost of a primary run.

\section{The Test Suite}
\label{app:findings}
Each test is a YAML file with an id, scope, status, description, \texttt{selector},
an \texttt{assertion} about future corpora, and an \texttt{evidence} block with the
measured prevalence and Wilson interval, the auditor's own expected precision, the
human labels and the ids it inspected.

Each baseline draws the most extreme \budget{} documents of its own signal,
such as the lowest confidences (0.667--0.671 here) or the largest duplicate clusters
(1{,}349--1{,}370{,}347), and never a lukewarm sample from the thresholded
slice, so every strategy sees the strongest version of the heuristic it stands for.
The thresholds quoted in \S\ref{sec:setup} are used for the overlap analysis, and we
check that every drawn document falls inside them.

Self-termination is likewise checked rather than assumed. Each run ends with
\texttt{stop\_reason} \texttt{end\_turn}, no error and its own written summary, after 67,
85 and 78 turns against a cap of 250, so the runs stopped because the model stopped,
not because the harness cut them off.

So are the inspection budgets. Every baseline sample holds exactly
300 documents, and each agent run's inspection counter is monotone, stays inside the
same allowance and ends at the number the results quote (197, 180, 173). The counter
is authoritative over the per-call field, because the frozen logs predate a fix in which each
call reported the rows it returned instead of the documents it newly charged, which
over-counted a re-read by 3, 8 and 10 calls respectively. The two now agree by
construction, and the framework self-test asserts it.

The release also carries a generated report, \texttt{REPORT.md}, which restates
every finding with its selector, the evidence the auditor gathered, its likely
false positives, documents to start from, static-slice overlap, the per-crawl
picture and the cost comparison. It is generated from the analysis artefacts, never written by hand, so it cannot
drift from the measurements, and it is what a curator
reads before deciding whether to act.

Findings compile into two kinds of test. \emph{Prevalence guards} fail if
a slice grows in a future release, and \emph{invariants} fail if a condition
occurs at all. A guard's bound, 1.2 times the measured prevalence, is calibrated for
corpus-wide evaluation. Over a narrow slice of crawls the same bound is easily
exceeded by ordinary variation (2 of the seven guards over the eight newest
snapshots, all seven over the single worst one), which is information about the crawl
rather than a regression. For the content farm, the guard (1.2 times 5.32\%) trips on
the crawls around the peak, roughly 2022-40 to 2023-23, and not on the two newest,
which is the guard working, since the farm far exceeds its corpus-wide share there. Both kinds are released, and a failing test is a live defect,
not a broken harness. Each kind carries what makes it retraceable, and the release check
enforces this. A guard ships the auditor's stated evidence, up to five document ids
to start from (all 32 resolve in the corpus) and the false positives it expects, while an
invariant has none of those by construction and instead ships the mechanism that
produces the violation.

Every finding retained after adjudication follows, as released. \texttt{accepted} marks a
finding the blinded labels support and whose adopted verdict is \texttt{approve},
and \texttt{candidate}
marks one kept in the suite without that support (here the \texttt{slo.*} farm).
\raggedbottom 
\providecommand{\findsel}[1]{\par\smallskip{\setlength{\fboxsep}{2pt}\colorbox{zebra}{\parbox{\dimexpr\linewidth-4pt}{\raggedright\ttfamily\scriptsize #1}}}\par}
\providecommand{\findkey}[1]{\textcolor{black!55}{#1}}
\providecommand{\findstats}[6]{\par\smallskip{\scriptsize\renewcommand{\arraystretch}{1.0}%
  \begin{tabular}{@{}l@{\hspace{6pt}}l@{}}
  \findkey{prevalence} & #1\% \enspace(#2 documents, exact)\\
  \findkey{defect rate} & #3 \enspace($n=#4$)\\
  \findkey{found in run} & #5\\
  \findkey{static overlap} & #6\%\\
  \end{tabular}}\par}
\providecommand{\findexc}[1]{\par\smallskip{\scriptsize\itshape\color{black!75}#1\par}}
\newtcolorbox{findingcard}[2]{enhanced, unbreakable, colback=white, colframe=rule,
  boxrule=0.4pt, arc=1.5pt, left=4pt, right=4pt, top=3pt, bottom=3pt,
  before skip=6pt, after skip=6pt,
  fontupper=\footnotesize, before upper={\rightskip=0pt plus 2.5em\relax\parfillskip=0pt plus 1fil\relax},
  fonttitle=\ttfamily\scriptsize\bfseries, coltitle=acc, colbacktitle=accsoft,
  title={#1\hfill#2}, toptitle=1.5pt, bottomtitle=1.5pt}
\begin{findingcard}{\texttt{adult\_spam\_farm\_wordsalad}}{\accepted}
A cluster of throwaway-domain adult/porn spam sites contribute documents that are not coherent Slovak prose but auto-generated word-salad: SEO tag lists, stitched keyword fragments, and repeated boilerplate phrases (e.g. "zadarmo čierne Lesbičky Sex Movies" injected dozens of times) built by a domain-spinning template. The corpus's own language\_score is near-perfect (often >0.99) and all three shipped LID tools agree on 'sk', yet the "documents" are not naturally authored Slovak text and provide no usable linguistic signal.

\findsel{selector: n\_bad\_words >= 3}

\findstats{0.387}{102{,}474}{0.84}{19}{rep2}{11}

\findexc{\emph{Excerpt withheld: every document in this slice matches the profanity list. See the released examples.}}
\end{findingcard}

\begin{findingcard}{\texttt{czech\_slovak\_code\_switched\_docs}}{\accepted}
A systematic subset of documents are Czech-language or heavily Czech/Slovak code-switched, not clean Slovak, yet are labelled language='slk' with high language\_score. This mostly comes from (a) Slovak e-commerce sites that paste Czech-language product descriptions/reviews verbatim from shared CZ/SK catalogs and (b) cz.pinterest.com / forum pages whose captions and comments mix Czech and Slovak. Because Czech and Slovak are highly mutually intelligible, the pipeline's own LID and all three shipped auxiliary LIDs (fasttext, langdetect, lingua) confidently but wrongly agree that many such documents are Slovak, and even when they agree the document is Czech, the corpus still ships it labelled slk.

\findsel{selector: fasttext=\textquotesingle{}cs\textquotesingle{} AND langdetect=\textquotesingle{}cs\textquotesingle{} AND lingua\_top=\textquotesingle{}cs\textquotesingle{}}

\findstats{0.167}{44{,}251}{0.81}{21}{rep2}{100}

\findexc{\texttt{\upshape\scriptsize <urn:uuid:9c1d3b82-06a4-49fd-81e3-c1c0e6ed864f>}: ``Žiadne produkty Bazénová chémia Napsali o nás Jsem spokojen zboži bylo rychle a v pořádku doručeno Diky 2019-02-12 Po objednání zboží, bylo dodáno na druhý den. Vše probě\dots''}
\end{findingcard}

\begin{findingcard}{\texttt{czech\_text\_mislabeled\_as\_slovak}}{\accepted}
A systematic slice of documents are substantially or entirely Czech-language text (product data, forum posts, cookie notices, thesis metadata, Pinterest boards) that the corpus nonetheless labels as Slovak (slk\_Latn). These typically come from bilingual CZ/SK commerce platforms, cross-posted forums, and Pinterest country mirrors that reuse Czech copy on .sk domains or vice versa.

\findsel{selector: fasttext=\textquotesingle{}cs\textquotesingle{} AND langdetect=\textquotesingle{}cs\textquotesingle{} AND lingua\_top=\textquotesingle{}cs\textquotesingle{} AND NOT (domain LIKE \textquotesingle{}\%.firebaseapp.com\textquotesingle{} OR domain LIKE \textquotesingle{}\%.web.app\textquotesingle{})}

\findstats{0.120}{31{,}795}{0.53}{17}{rep1}{100}

\findexc{\texttt{\upshape\scriptsize <urn:uuid:9c1d3b82-06a4-49fd-81e3-c1c0e6ed864f>}: ``Žiadne produkty Bazénová chémia Napsali o nás Jsem spokojen zboži bylo rychle a v pořádku doručeno Diky 2019-02-12 Po objednání zboží, bylo dodáno na druhý den. Vše probě\dots''}
\end{findingcard}

\begin{findingcard}{\texttt{firebase\_content\_farm\_spam}}{\accepted}
A large-scale, machine-generated content-farm/SEO-spam network is hosted on randomly-named subdomains of Google's free Firebase hosting (*.firebaseapp.com, *.web.app). Each subdomain name is a Slovak or Scandinavian finance/get-rich-quick keyword (e.g. "akozbohatnut" = how to get rich, "investerarpengar" = Swedish "invest money", "pracovnemiesta" = job vacancies, "danove" = taxes, "nakupzlata" = buy gold) plus a random letter suffix, and each hosts thousands of pages of incoherent, multi-topic text stitched together from scraped fragments (finance/crypto glossary snippets, news snippets, unrelated Q\&A, sometimes switching mid-document into Czech, Bosnian/Serbian, or Slovenian). The pages are not authored prose and are not really "Slovak documents" in the intended sense, despite passing the language filter with high average confidence (0.98) and evading near-duplicate dedup entirely (avg minhash\_cluster\_size $\approx$ 1, since every domain/page uses a unique random suffix/URL).

\findsel{selector: domain LIKE \textquotesingle{}\%.firebaseapp.com\textquotesingle{} OR domain LIKE \textquotesingle{}\%.web.app\textquotesingle{}}

\findstats{5.321}{1{,}408{,}437}{0.74}{19}{main, rep1, rep2}{13}

\findexc{\texttt{\upshape\scriptsize <urn:uuid:4ab0bcc3-6437-477c-a3bc-00cdc268b05a>}: ``Kľúčenka na nákupy mincí Skvelá kľúčenka pre menšie deti mini batôžtek na sladkosti alebo pre väčšie slečny vrecúško na drobnosti mince a pod. Javascript není prohlížečem\dots''}
\end{findingcard}

\begin{findingcard}{\texttt{porn\_doorway\_spam\_network}}{\accepted}
A second, distinct content-farm/doorway-page spam network -- unrelated to the firebaseapp/web.app finance farm -- serves auto-generated pornography aggregator pages under \textasciitilde{}60+ unrelated-looking domains with generic personal/company names and varied TLDs (adv-moto-girl.bike, baazzaa.com, agencerepublique.net/.me, aaronpowell.me, bananapepper.fr, americanislandsregionalcenter.us, etc.). Pages are keyword-stuffed concatenations of pornographic video-tag phrases machine-translated/rendered into Slovak, not coherent prose, and repeat the same tag vocabulary across thousands of near-identical URLs per domain.

\findsel{selector: n\_bad\_words > 5}

\findstats{0.198}{52{,}354}{1.00}{20}{rep1}{10}

\findexc{\emph{Excerpt withheld: every document in this slice matches the profanity list. See the released examples.}}
\end{findingcard}

\begin{findingcard}{\texttt{porn\_spam\_keyword\_farm}}{\accepted}
A second, separate spam network of thousands of randomly-named domains (English dictionary-word-like names + generic gTLDs/ccTLDs such as .com/.me/.fr/.bike/.online/.info/.nl/.it/.us/.br) hosts keyword-stuffed, incoherent "word salad" adult/porn doorway pages that are grammatically broken pastiches of scraped Slovak sentence fragments and tag lists, not genuine documents. Many share a literal CMS template header "Posted by Admin | in <Month> <Day>, 2020 |".

\findsel{selector: n\_bad\_words >= 3 AND domain NOT LIKE \textquotesingle{}\%.sk\textquotesingle{} AND domain NOT LIKE \textquotesingle{}\%.cz\textquotesingle{}}

\findstats{0.359}{94{,}946}{0.83}{12}{main}{11}

\findexc{\emph{Excerpt withheld: every document in this slice matches the profanity list. See the released examples.}}
\end{findingcard}

\begin{findingcard}{\texttt{slo\_prefix\_machine\_translated\_farm}}{\candidate}
A third distinct content-farm network: hundreds of unrelated-looking, seemingly hijacked/repurposed domains (e.g. slo.andhramedicalcollegevizag.com, slo.istanbulmedicine.com, slo.kelinvestgroup.com, slo.orsitaning.ru) all serve the same machine-translated content mirror under a 'slo.' (Slovak) subdomain. Content is overwhelmingly pseudo-medical/health/lifestyle articles whose URL slugs are transliterated Russian (e.g. 'diagnostika-raka-jaichek-u-muzhchin', 'kod-po-mkb-mehanicheskaja-zheltuha'), indicating the same Russian-origin article set has been bulk machine-translated into Slovak (and presumably many other languages via sibling subdomain prefixes) and republished across this domain farm. Unlike the firebaseapp.com spam, the translated text is fluent and grammatically coherent single-topic prose, so it reads as legitimate native Slovak and would not be caught by quality heuristics, but it is not organically authored Slovak content and the same articles are repeated near-verbatim across dozens of domains.

\findsel{selector: domain LIKE \textquotesingle{}slo.\%\textquotesingle{}}

\findstats{0.279}{73{,}835}{0.07}{14}{rep1}{0}

\findexc{\texttt{\upshape\scriptsize <urn:uuid:31d250de-9636-48d5-997e-a27d6aafc2db>}: ``Ruský biele kuracie plemeno, popis chovu Video: vajcia kuracie chovy obsah Video: Breed kurčatá - talianska, Menorca, Faverolles Kurča, Kuchinskaya výročie, Orpington Sli\dots''}
\end{findingcard}

\newpage
\section{Validation Details and Annotation Guidelines}
\label{app:annotation}
\paragraph{Labelling passes.} The first pass labelled the documents in their shuffled order
and stopped after the first \annpassone{}. After the criterion for junk was clarified, a
second pass judged \annpasstwo{} documents under it: \annrecodelisted{} documents had been
listed for re-coding, and the second pass covered 54 of them and 84 others, \annpasstwonew{}
of the \annpasstwo{} beyond the first pass's prefix. The other 117 listed documents keep
their first-pass labels. The union is \annlabelled{} labels, and documents neither pass
reached were dropped.

\paragraph{Corrected adjudication rows.} \textsc{Random} and \textsc{LowConf} proposed the
same English-text selector, and our tabulation merged the two under \textsc{LowConf}'s name
and lost the 7 labels drawn under \textsc{Random}'s, so both rows were first marked
\texttt{unsure} for want of labels. With the labels attributed by selector, the slice comes
out at .57 ($n=7$), and we changed both rows to \texttt{approve} by the stated rule, at the
author's direction and without a new deliberation.

\paragraph{Profanity-ranked sample.} The slice $\texttt{n\_bad\_words} \geq 3$
(\profslicedocs{} documents) is byte-identical to one G04 selector and contains
\profcoversall{} G04 selectors entirely. Among the \profbudget{} most profane documents in
the text subsample (\texttt{n\_bad\_words} \profbwmin--\profbwmax), all G04 by definition,
\profoutside{} belong to any other agent finding, so a profanity-ranked audit would read
only G04 documents.

\paragraph{Guidelines.} The guidelines given to the annotator, verbatim, in their final
version. The paragraph beginning ``Readability is not the test'' was added between the
two passes, when the junk criterion was clarified; the \annpassone{} first-pass labels
were given without it.
\par
\begin{artefactbox}{Annotation guidelines (verbatim, final version)}
{\footnotesize\begin{verbatim}
# Slovak FineWeb2 audit: annotation
# guidelines

You will see documents drawn from a
corpus that a curation pipeline labelled as
**Slovak** and shipped as pretraining data.
Documents are shown in random order; the
sampling method and the reason a document
was selected are deliberately hidden from
you.

For each document give **two** labels.

## 1. Language

* `sk`      Slovak. Normal Slovak text.
            Informal, diacritic-free or
            misspelled Slovak still counts
            as Slovak.
* `sk-pred` Predominantly Slovak: mostly
            Slovak, but with a non-trivial
            amount of another language
            (e.g. a Czech or English block).
* `cs`      Czech. The document is
            essentially Czech.
* `mix`     Mixed Slovak/Czech:
            substantial amounts of both,
            neither clearly dominant.
* `other`   Some other language, or
            genuinely multilingual beyond
            sk/cs.
* `?`       Uncertain.

## 2. Usefulness as Slovak corpus text

* `good`    Useful natural text: coherent
            prose a person wrote (or a
            decent translation) that you
            would keep in a Slovak corpus.
* `noisy`   Usable but noisy: real
            content mixed with menus,
            repetition, formatting debris,
            or awkward machine translation.
* `junk`    Boilerplate, template,
            navigation, listing spam,
            keyword salad, incoherent
            machine-generated text, or
            otherwise not usable.
* `?`       Uncertain.

Readability is not the test. Fluent,
correctly spelled Slovak is still `junk`
when the page exists to place links,
keywords or advertising rather than to
say something: doorway pages, tag and
category listings, autogenerated product,
crypto or gambling copy. `noisy` is for
real content with debris around it, not
for debris that happens to be readable.

Judge only what you see. If the excerpt
is truncated, judge the excerpt.
\end{verbatim}}
\egroup 
\end{artefactbox}

\end{document}